\documentclass[twoside]{article}
\usepackage[accepted]{aistats2014}
\usepackage{color}
\usepackage{graphicx}
\usepackage{amsmath}
\usepackage{amssymb}
\usepackage{multirow}
\usepackage[authoryear]{natbib}

\usepackage{hyperref}

\setcitestyle{round}

\newcommand{\corp}[1]{\textsl{#1}}
\newcommand{\Enhance}{\corp{Enhance}}
\newcommand{\An}{\corp{Astrometry.net}}
\newcommand{\flickr}{\corp{flickr}}
\newcommand{\foreign}[1]{\textit{#1}}

\newcommand{\red}[1]{#1}
\newcommand{\figref}[1]{Figure~\ref{#1}}
\newcommand{\thou}{,\!000}

\graphicspath{{jpeg-figs/}{jpeg-half-figs/}{example-figs/}}

\newlength{\twocolfigbottomspace}
\newlength{\figendspace}
\newcommand{\twocolfigbottom}{\vspace{\twocolfigbottomspace}}
\newcommand{\figbottom}{\vspace{\figendspace}}

\newcommand{\codevspace}{\vspace{-6pt}}

\newcommand{\codeminipage}{%
\begin{minipage}[l]{6.5cm}
\codevspace
\begin{eqnarray*}
\mbox{For}~i&=&1,\dots,N,~\mbox{do}:
\nonumber \\
\tilde{c} &:=& c|_{m_i}
\nonumber \\
\tilde{v} &:=& v|_{m_i}
\nonumber \\
\sigma &:=& \mbox{\tt argsort} (\tilde{c})
\nonumber \\
\tau &:=& \mbox{\tt argsort} (d_i)
\nonumber \\
\rho &:=& \mbox{\tt argsort} \left( \frac{\tilde{v} * \sigma^{-1} + \tau^{-1}}{\tilde{v}+1} \right)
\nonumber \\
\tilde{c} &:=& \tilde{c}\left(\sigma\left(\rho^{-1}\right)\right)
\nonumber \\
\tilde{v} &:=& \tilde{v} + 1
\nonumber \\
 c|_{m_i} &:=& \tilde{c}
\nonumber \\
 v|_{m_i} &:=& \tilde{v}
\end{eqnarray*}
\end{minipage}
}

\newcommand{\commentsminipage}{%
\begin{minipage}[r]{8cm}%{7.5cm}
\codevspace
{\footnotesize\em

\bigskip
\smallskip

loop over data images
\bigskip

extract sub-vectors according to the mask $m_i$ of current data image $d_i$\medskip

compute pixel rank vectors  $\sigma^{-1}, \tau^{-1}$ of consensus subimage and of data image and compute weighted average $\rho^{-1}$\medskip

compute weighted average of pixel ranks (multiplication and division are understood element-wise) and argsort\medskip

permute pixels of the consensus subimage such that their ranking agrees with the
  consensus pixel rank vector $\rho^{-1}$\smallskip

increment entries of voting subvector\smallskip

fill in updated subimage and votes into consensus image $c$ and voting vector $v$
}
\end{minipage}%
}

\newcommand{\codefigure}{%
\begin{figure}[t!]
\twocolumn[%
  \codeminipage
  \vline ~~
  \commentsminipage
  \caption{Proposed \Enhance\ algorithm.
    \label{fig:code}
  }
  \twocolfigbottom
]
\figbottom
\end{figure}
}

\newcounter{thumbnail}
\newcommand{\spc}{\hspace{0.01\textwidth}}
\newcommand{\tlab}[1]{\raisebox{1ex}{\makebox[0in][r]{\textbf{\textcolor{green}{(\stepcounter{thumbnail}\alph{thumbnail})#1}}}}}
\newcommand{\solved}{\tlab{$\,\ast$}}
\newlength{\imh}
\newcommand{\img}[1]{\includegraphics[height=\imh]{#1}}
\newcommand{\uimg}[1]{\img{#1}\tlab{}}
\newcommand{\simg}[1]{\img{#1}\solved}

\newcommand{\examplefigurecontents}{%
\uimg{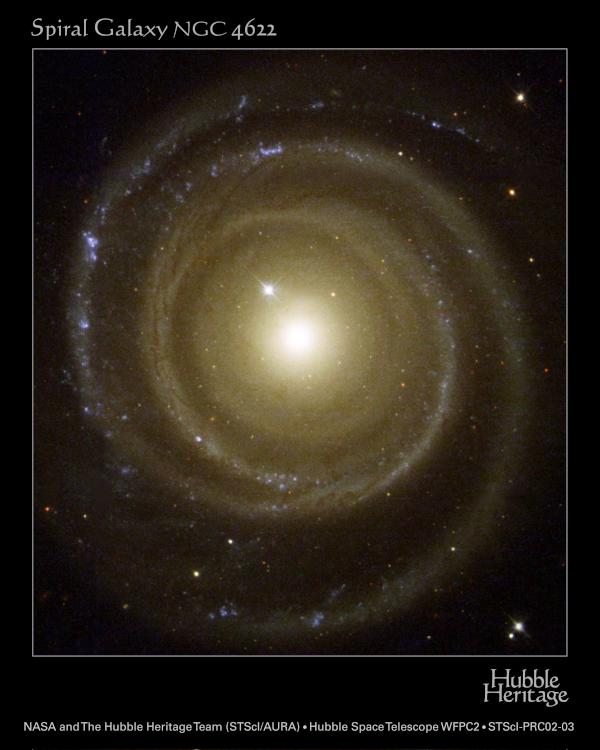}\spc%
\simg{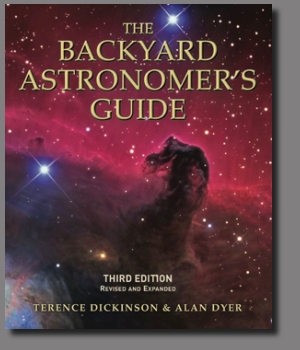}\spc%
\uimg{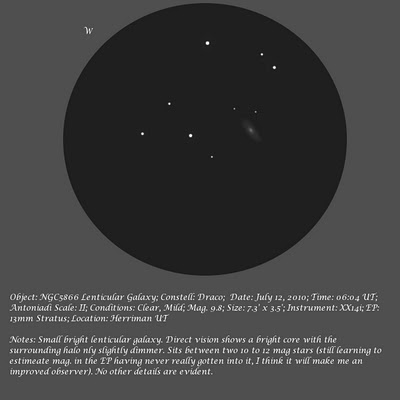}\spc%
\simg{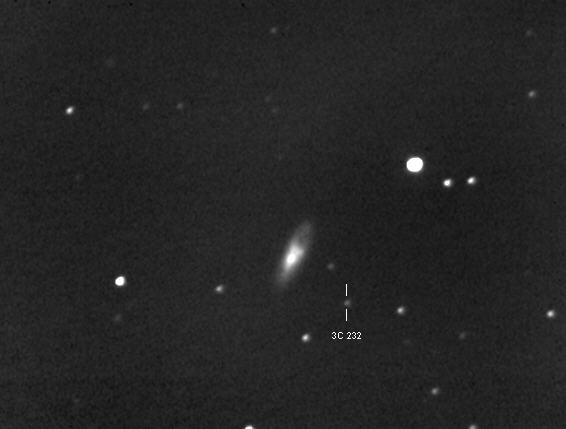}\spc%
\uimg{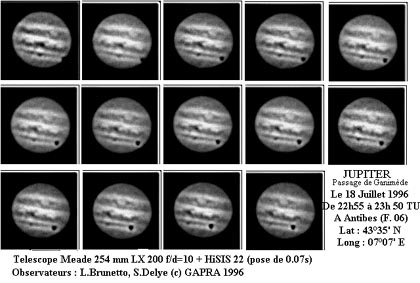}%

\vspace{3pt}%
\setlength{\imh}{0.13\textwidth}
\uimg{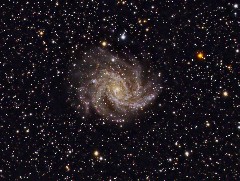}\spc%
\uimg{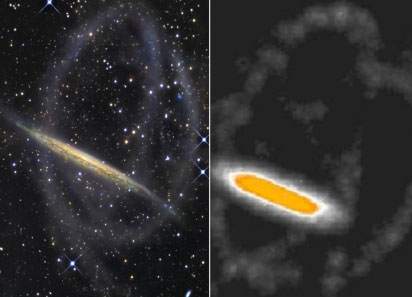}\spc%
\simg{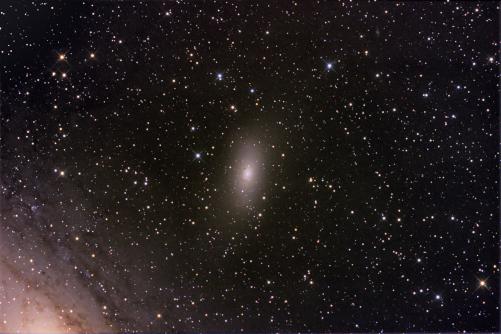}\spc%
\simg{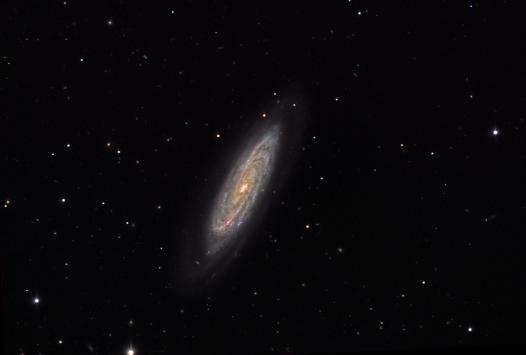}\spc%
\simg{01db7f4e9c452cbddfeaf49aadfa3109}%
\caption{Example images found through our Web search for images of
  NGC~5907.  Images that were successfully calibrated by \An\ are
  marked with $\ast$.  Notice that for this rather distinctive search
  term, most images are actually astronomical in nature but relatively
  few are actually images of NGC~5907---only \textsl{(g)} and
  \textsl{(j)} here, and roughly 450 of the 2000 images resulting from
  our Web search.
  \label{fig:examples}}
}

\newcommand{\examplefigure}{%
\begin{figure}[t!]
\twocolumn[
\examplefigurecontents
\twocolfigbottom
]
\figbottom
\end{figure}
}

\newcommand{\theonelinefig}{
  \renewcommand{\arraystretch}{0}
  \newlength{\figw}
  \newlength{\bigw}
  \newlength{\rowspace}
  \setlength{\rowspace}{2pt}
  \setlength{\figw}{0.09\textwidth}
  \setlength{\bigw}{4\figw}
  \addtolength{\bigw}{3\rowspace}
  \newlength{\upp}
  \setlength{\upp}{\figw}
  \begin{tabular}{c@{\hspace{3pt}}c@{\hspace{3pt}}c@{\hspace{3pt}}c}
    Images &
    Histogram Eq. &
    Hist.-Eq.~weighted average &
    \Enhance\ result \\[\rowspace]
    \includegraphics[width=\figw]{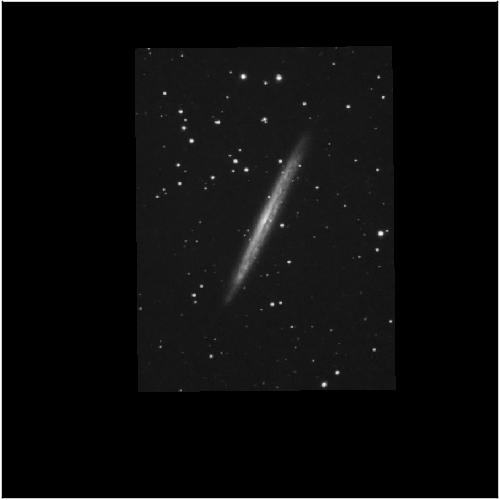} &
    \includegraphics[width=\figw]{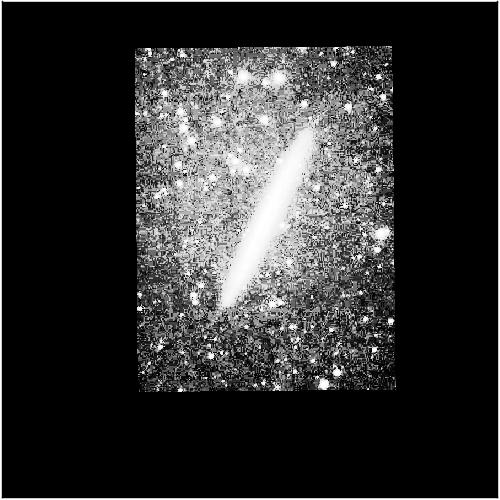} &
    \multirow{4}{*}[\upp]{\includegraphics[width=\bigw]{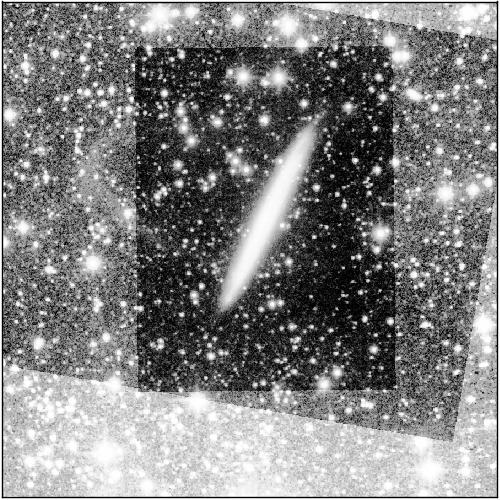}} &
    \multirow{4}{*}[\upp]{\includegraphics[width=\bigw]{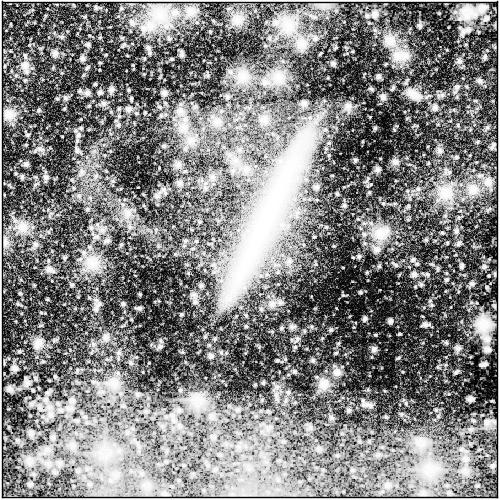}}
    \\[\rowspace]
    \includegraphics[width=\figw]{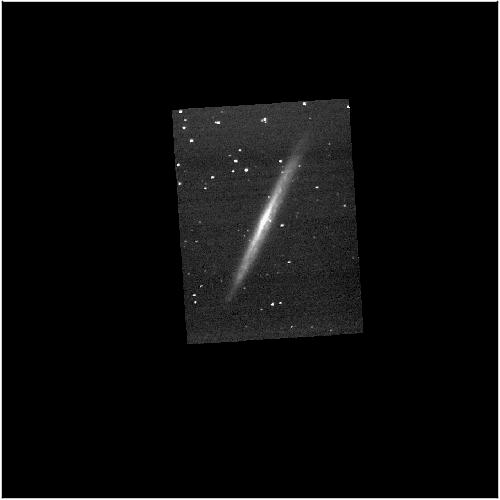} &
    \includegraphics[width=\figw]{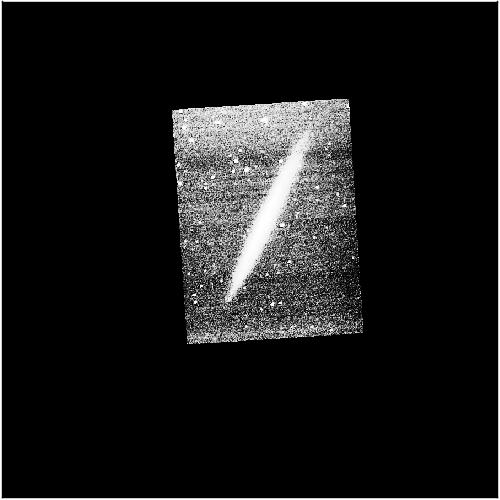} &
    \\[\rowspace]
    \includegraphics[width=\figw]{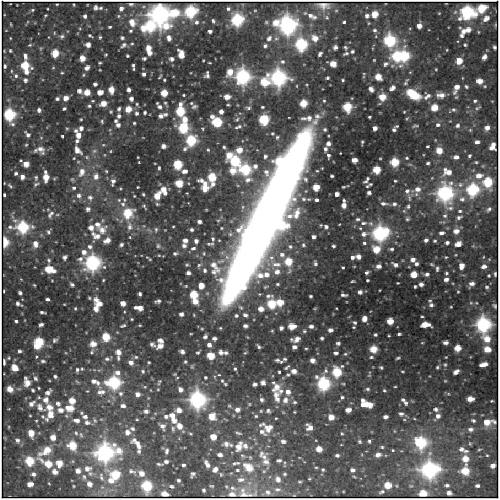} &
    \includegraphics[width=\figw]{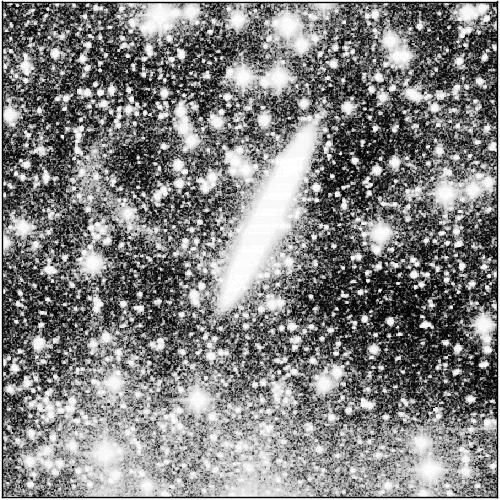} &
    &
    \\[\rowspace]
    \includegraphics[width=\figw]{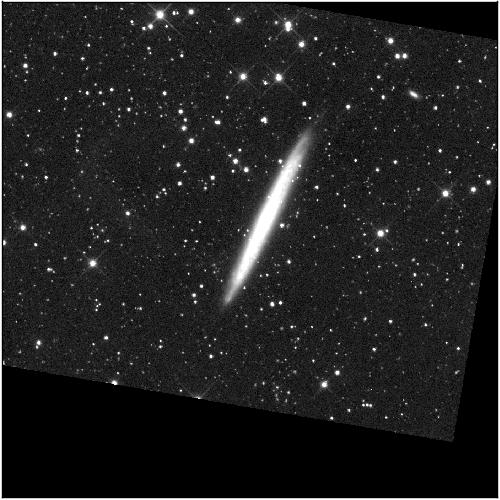} &
    \includegraphics[width=\figw]{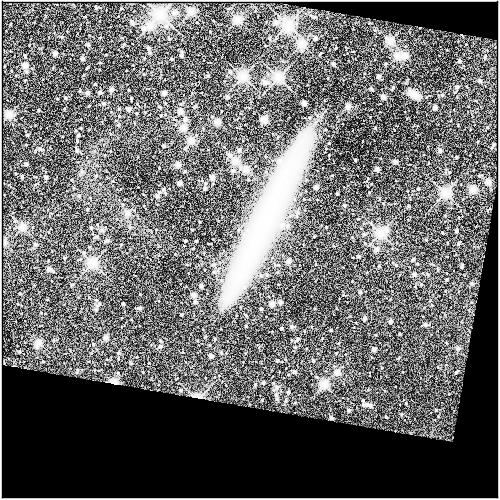} &
    &
  \end{tabular}
}

\newcommand{\histeqfigure}{%
\begin{figure}[t!]
  \twocolumn[
    \centering
    \theonelinefig
    \caption{Four example images from the NGC~5907 collection, chosen to
    show very different dynamic range and pre-processing.
    \textbf{Left column:} Input images (resampled to the common pixel
    grid).  The top two images have very low saturation (very few
    pixels with the maximum value); the bottom two images have high
    saturation.  \textbf{Middle column:} Histogram-equalized input
    images.  \textbf{Top-right:} The histogram-equalized average of
    the input images; image borders are clearly visible and the
    resulting image does not clearly show more dynamic range than the
    inputs.  \textbf{Bottom-right:} Result of running the
    \Enhance\ algorithm.  The input image edges are less prominent,
    and the looping stream feature as well as the details in the core
    of the galaxy are visible due to the enhanced dynamic range.
    \label{fig:four}}
    \twocolfigbottom
  ]
  \figbottom
\end{figure}
}

  \renewcommand{\arraystretch}{1}

\newcommand{\ngcresultsfigcontents}{%
    \newlength{\smallha}
    \setlength{\smallha}{0.095\textwidth}
    \newlength{\smallhb}
    \setlength{\smallhb}{0.086\textwidth}
    \centering
    \begin{tabular}{c@{\hspace{3pt}}c@{\hspace{3pt}}c}
      \multirow{2}{0.25\textwidth}[0.075\textwidth]{%
        \includegraphics[width=0.25\textwidth]{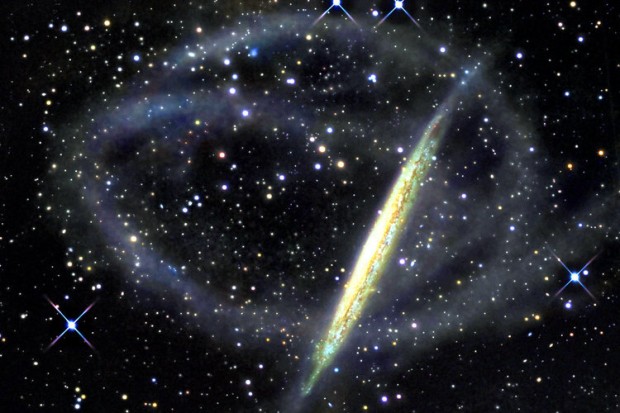}
      }
      &
      \makebox[0.25\textwidth][l]{%
        \includegraphics[height=\smallha]{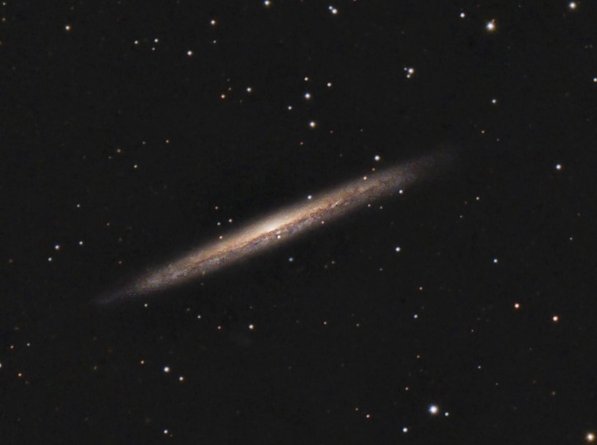}
        \includegraphics[height=\smallha]{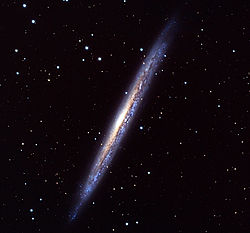}
        \includegraphics[height=\smallha]{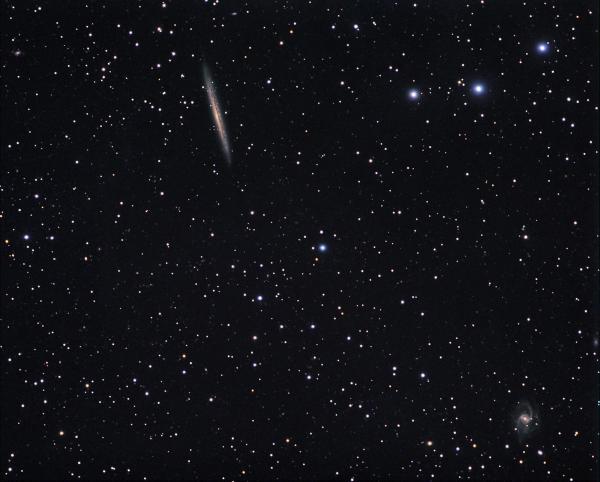}
        \includegraphics[height=\smallha]{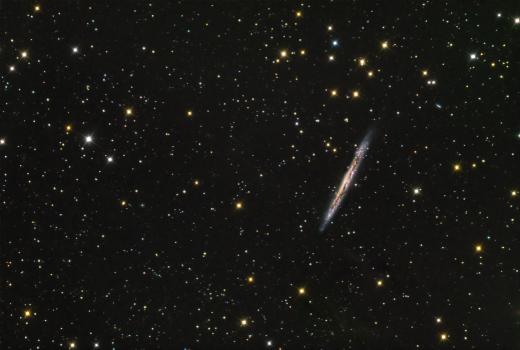}
      }
      &
      \\
      &
      \makebox[0.25\textwidth][l]{%
        \includegraphics[height=\smallhb]{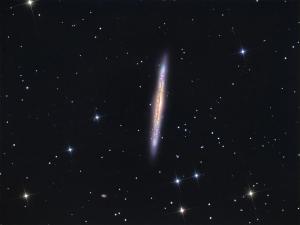}
        \includegraphics[height=\smallhb]{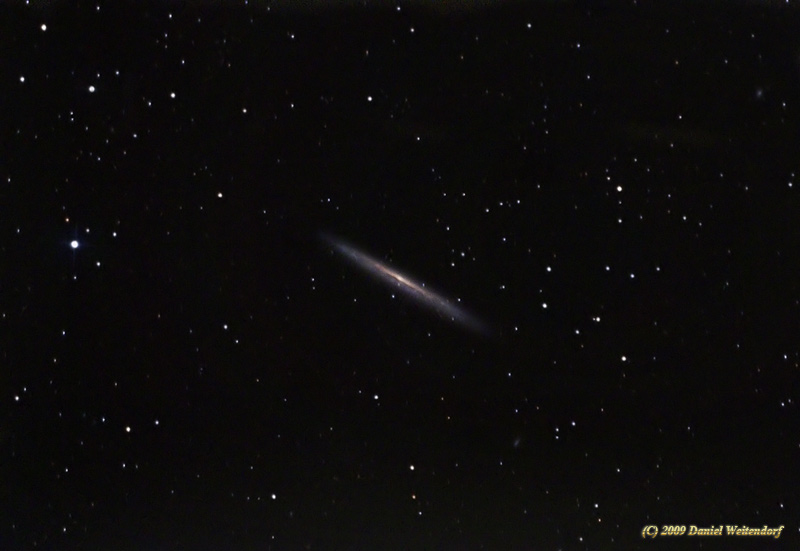}
        \includegraphics[height=\smallhb]{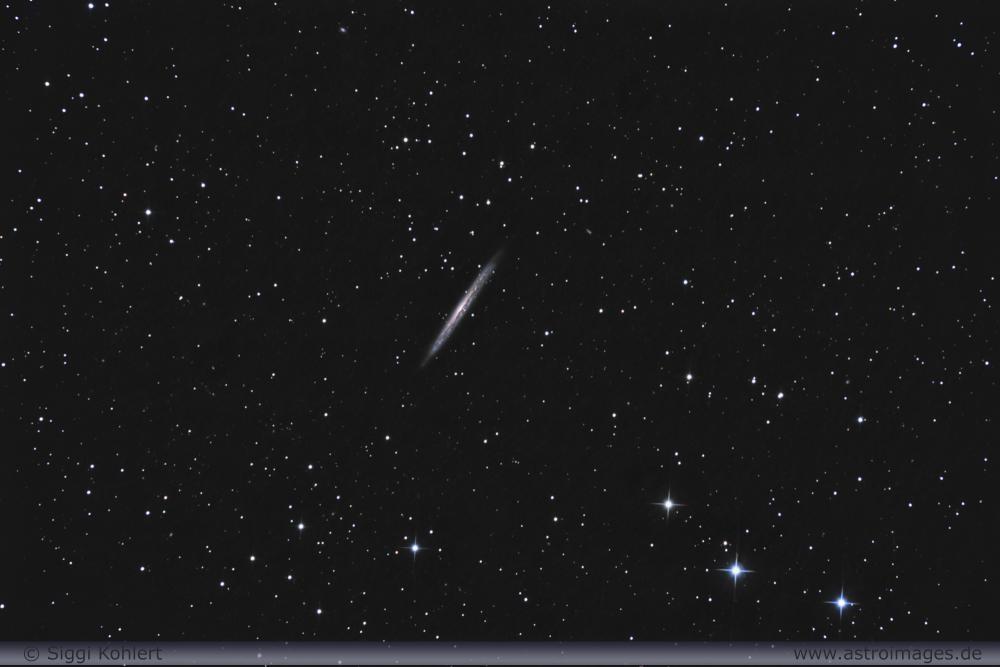}
        \includegraphics[height=\smallhb]{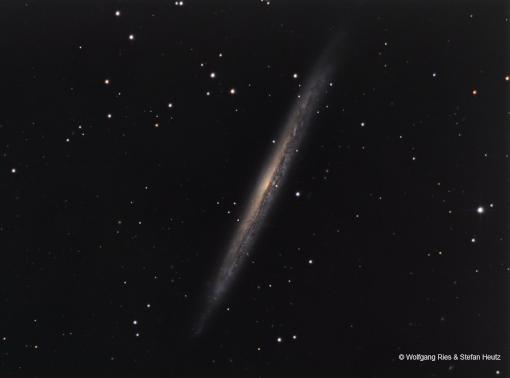}
      }
      \\
      \includegraphics[width=0.25\textwidth]{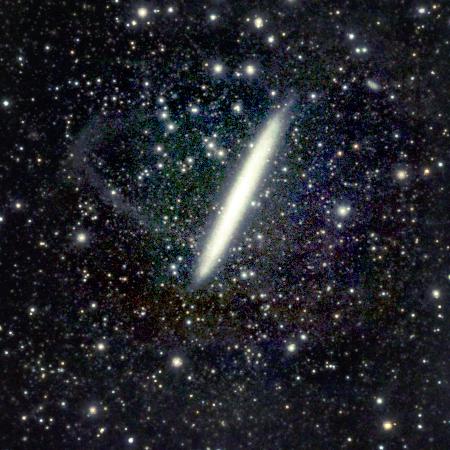} &
      \includegraphics[width=0.25\textwidth]{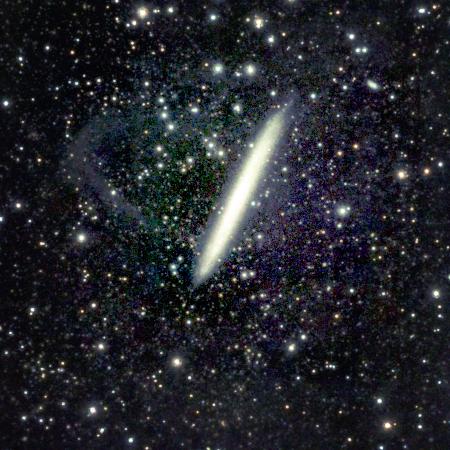} &
      \includegraphics[width=0.25\textwidth]{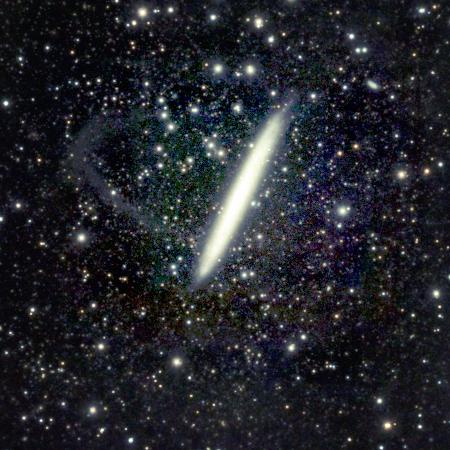}
      \\
    \end{tabular}
\caption{\emph{Top left}: deep image from \citet{dmd2008}, showing a faint
  stellar stream. This is a visualization of the deepest image ever
  taken of NGC~5907. \emph{Top right}: example images from our set
  of 298 images, obtained by a Web search for NGC~5907 (see
  \figurename~\ref{fig:examples}), and manually removing images that
  show the stellar stream visible in the deep image. \emph{Bottom, left to
    right}: results obtained by our algorithm, tone-mapped to have the
  same histogram as the deep image: after one run through
  the image set; after one run using a different permutation of the image
  set; and ten runs through the image set.  While the algorithm
  is not strictly invariant to image reordering, the results are
  visually quite similar. Also, it is noteworthy that one pass through
  the dataset gives very good results, showing the faint stellar
  stream visible in the deep image, which required over 11 hours of exposure
  time on a 0.5-meter telescope.\label{fig:5907}}
}

\newcommand{\ngcresultsfigure}{%
\begin{figure}[t!]
  \twocolumn[
    \ngcresultsfigcontents
    \twocolfigbottom
  ]
\figbottom
\end{figure}
}

\newcommand{\mfiftyonecontents}{%
  \centering
    \includegraphics[width=0.2\textwidth]{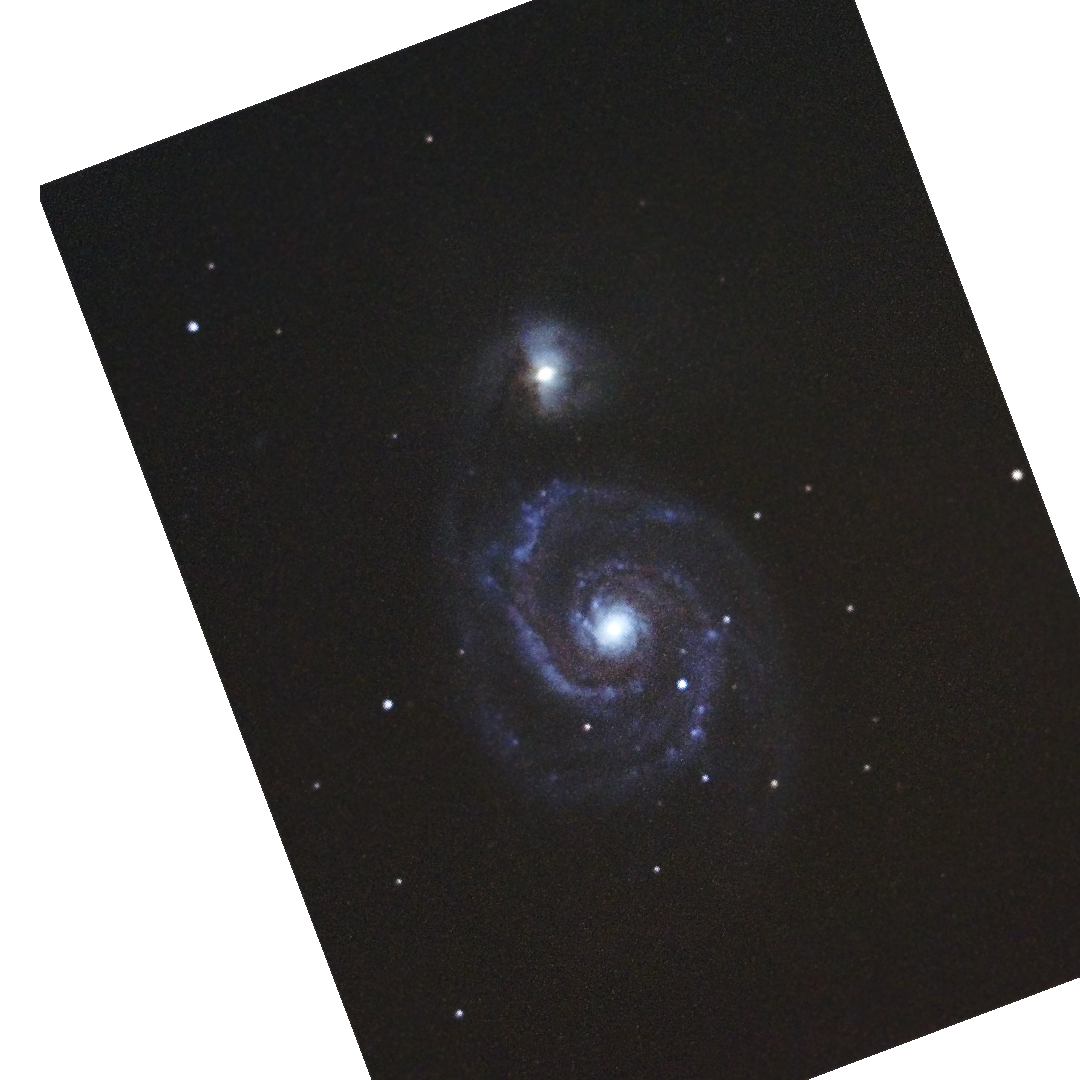}\hspace{3pt}%
    \includegraphics[width=0.2\textwidth]{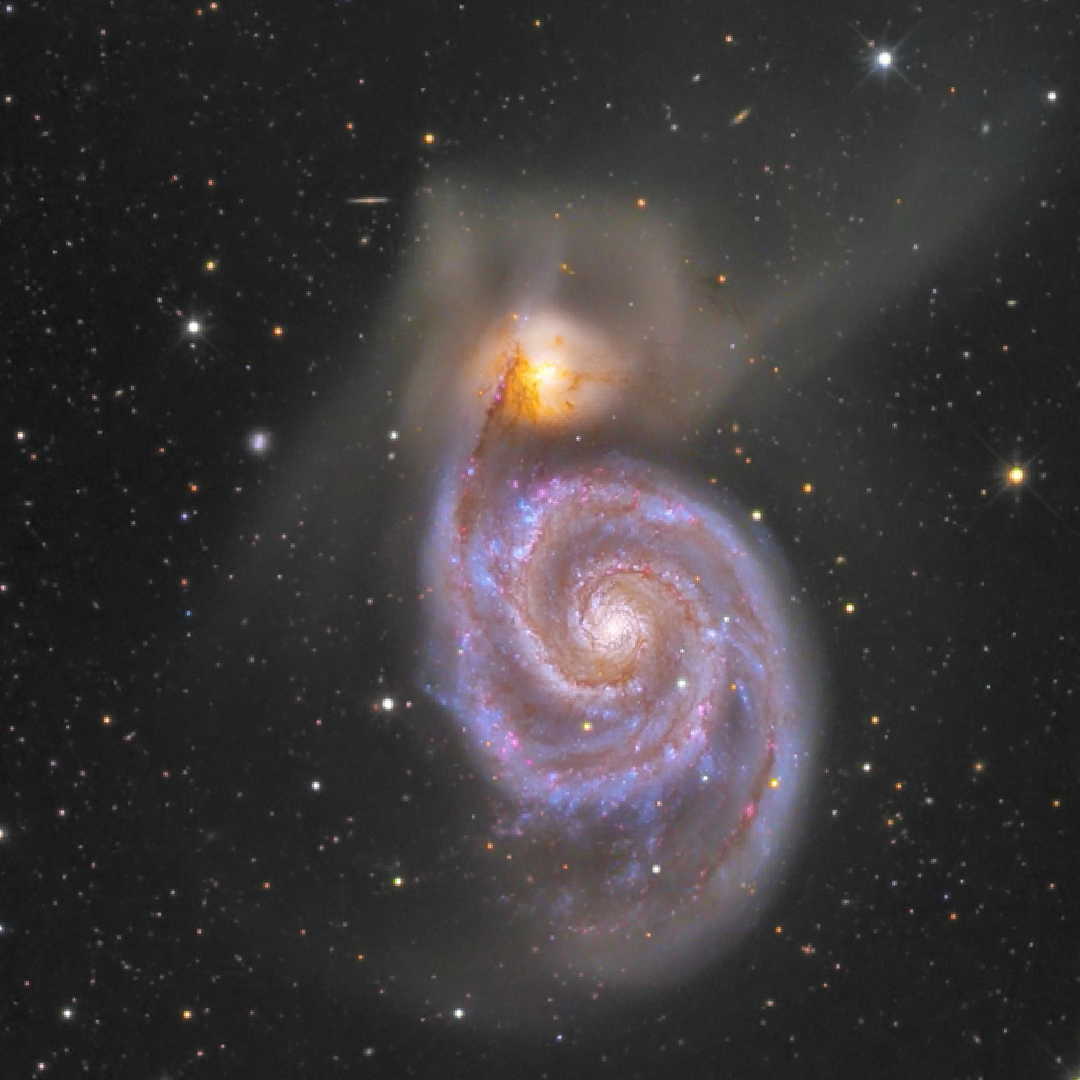}\hspace{3pt}%
    \includegraphics[width=0.2\textwidth]{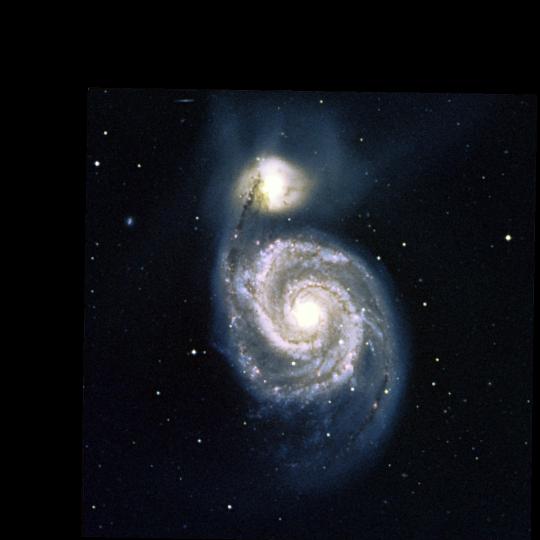}\hspace{3pt}%
    \includegraphics[width=0.2\textwidth]{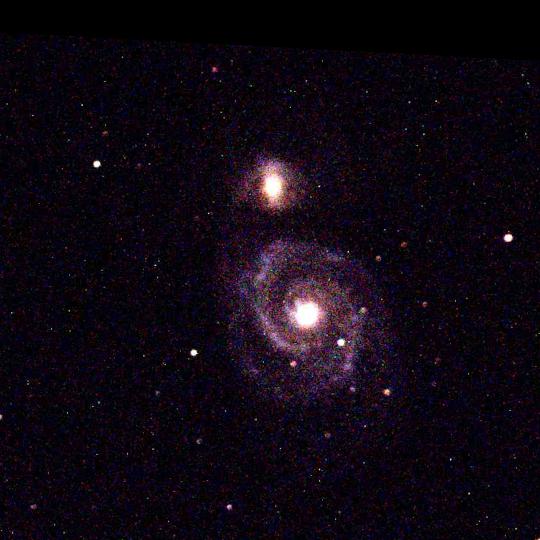}%
    \\[3pt]
    \includegraphics[width=0.4\textwidth]{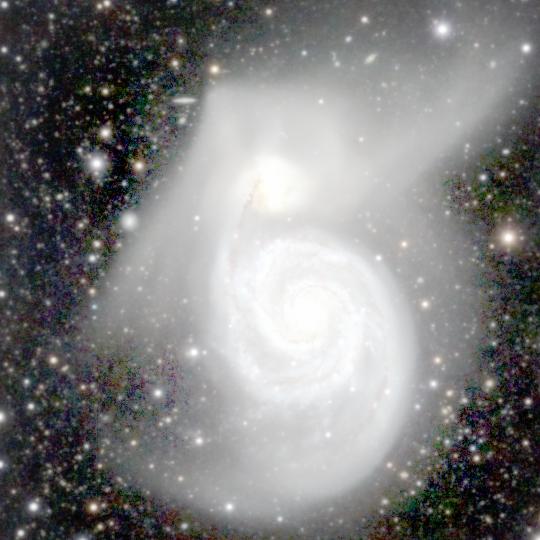}\hspace{9pt}%
    \includegraphics[width=0.4\textwidth]{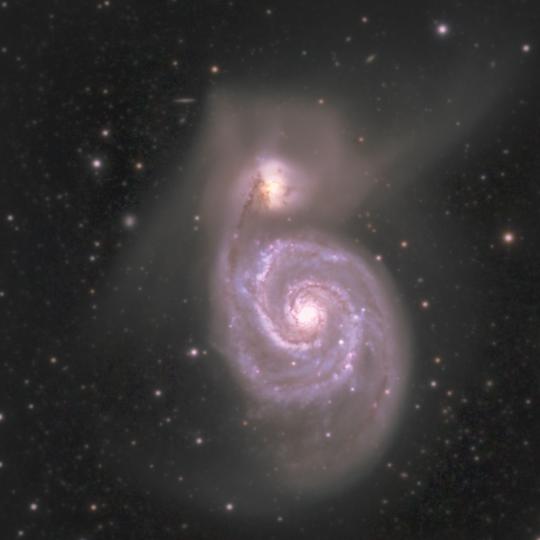}%
  \caption{Example images from the M51 collection (resampled to the
    reference grid), and the result of the \Enhance\ algorithm acting
    on all 2066 images in the collection; \emph{Bottom left:}
    Histogram-equalized consensus image; \emph{Bottom right:} Consensus
    image, tone-mapped to the second image in the top row.  Notice the
    faint extended structures around both galaxies: this is the debris
    resulting from their strong gravitational interaction.
    \label{fig:m51}}%
}

\newcommand{\mfiftyonefigure}{%
\begin{figure}[t]
\twocolumn[
  \mfiftyonecontents
  \twocolfigbottom
]
\figbottom
\end{figure}%
}

\begin{document}

\runningauthor{Lang, Hogg, \& Sch\"olkopf}

\twocolumn[
\aistatstitle{Towards building a Crowd-Sourced Sky Map}
\aistatsauthor{Dustin Lang \\
McWilliams Center for Cosmology \\
Carnegie Mellon University \\
5000 Forbes Ave, \\ Pittsburgh, PA 15213 \\
\texttt{dstn@cmu.edu} \\
\And
David W. Hogg \\
Center for Cosmology and \\ Particle Physics \\
New York University \\
4 Washington Place, \\ New York, NY 10003 \\
\texttt{david.hogg@nyu.edu} \\
\vspace{1em} 
\And
Bernhard Sch\"olkopf \\
Max Planck Institute for \\ Intelligent Systems \\
Spemannstrasse 38, \\ 72076 T\"ubingen, Germany \\
\texttt{bs@tuebingen.mpg.de} \\
}
]
\begin{abstract}
  We describe a system that builds a high dynamic-range and wide-angle
  image of the night sky by combining a large set of input images. The
  method makes use of pixel-rank information in the individual input
  images to improve a ``consensus'' pixel rank in the combined image.
  Because it only makes use of ranks and the complexity of the
  algorithm is linear in the number of images, the method is useful
  for large sets of uncalibrated images that might have undergone
  unknown non-linear tone mapping transformations for visualization or
  aesthetic reasons.  We apply the method to images of the night sky
  (of unknown provenance) discovered on the Web. The method permits
  discovery of astronomical objects or features that are not visible
  in any of the input images taken individually. More importantly,
  however, it permits scientific exploitation of a huge source of
  astronomical images that would not be available to astronomical
  research without our automatic system.
\end{abstract}

\section{Introduction}

The Web contains millions of astronomical images in human-viewable
formats (eg, jpeg)---the ``Astrophotography'' group on \flickr\ alone
has over $68\thou$ images---and these images contain scientifically
valuable information about the night sky.  Some of these images are
night-time snapshots, some are carefully rendered visualizations using
many images taken by high-quality backyard telescopes.  Either
way, these images are difficult to use for scientific purposes because
they have unknown provenance, and images rendered for human visual
consumption are often processed with non-trivial non-linear (and
sometimes non-local) transformations.  Such transformations are often
tone mappings applied to the intensity values (e.g., the gamma
transform that is applied to all jpeg images).  We also often know
nothing (\foreign{a priori}) about the point-spread function or
vignetting or noise or distortions.

Nonetheless, it remains a holy grail of ``internet astrophysics'' to
learn everything about the night sky that can be learned from the
images on the Web; in principle these images may contain a huge amount
of information about rare transients, variable stars, and high
proper-motion objects.  One example of a successful exploitation of
Web imaging is the determination of the gravitational orbit of Comet
17P/Holmes using imaging found by Web search \citep{lang2011}; another
is the determination of the impact trajectory of the Chelyabinsk
meteoroid from security-camera and other footage \citep{zuluaga2013}.
An interesting new direction---explored here---is to combine the
imaging into an all-sky ``consensus image'' that ideally contains
information about the static sky from all images ever taken.  This
project could take the form of a massive citizen-science project or as
a Web-scraping project.

One of the most important capabilities of a consensus image would be
its great sensitivity to faint, extended features such as the
outskirts of nearby galaxies, nebulae, and---as we explore here---the
faint streams of debris that result when two galaxies interact.  For
extended features, the combined imaging from many small telescopes can
be competitive with large telescopes.  Indeed, many of the faintest
features known in nearby galaxies were found with telescopes 16 to 50
cm in diameter \citep{dmd2008, dmd2009, dmd2010}.

%; see \figref{fig:5907}.

%  at faint isophotal levels.
% Because intensity is a conserved quantity, the surface-brightness
% sensitivity of a sky-photon-limited telescope is not a strong function
% of its aperture; the combined imaging from many small-telescope
% observations might contain a lot of information about faint emission.
% Indeed, many of the faintest features known in nearby galaxies were
% found with telescopes with small apertures of 16 to 50 cm diameter
% \citep{dmd2008, dmd2009, dmd2010}.

The key problem we face is extraction of scientifically valuable
information from the images without detailed knowledge of image
provenance or processing.  In particular, we want to combine the
information from many images to produce highly informative, high
dynamic-range images that go much deeper than any individual
contributing image, and we want to do this without having to infer the
(possibly highly nonlinear) processing that has been applied to each
individual image, each of which has been wrecked in its own loving way
by its creator.

The problem we are tackling is different than related problems because
we are truly photon-limited: we want to increase the effective
exposure time (sensitivity) of our consensus image by combining the
exposure time of a large number of independent input images.  In
astronomical images, it is nearly always that case that if we had
additional exposure time, we would detect more (faint) objects, or
more clearly resolve faint extended features.  In typical \emph{high
  dynamic range} imaging problems known from computer vision, in
contrast, the issue is to capture and represent (perhaps artistically)
an increased range of contrasts; it is assumed that it is easy to
\emph{capture} the image with the longest exposure time in a single
exposure, and typically only a few exposure times are required to
capture a satisfactory dynamic range.  In \emph{lucky imaging}
problems \citep{law2006, lucky}, the goal is typically to achieve the
highest possible \emph{resolution}; sensitivity or signal-to-noise is
secondary.
In \emph{panorama stitching} \citep{brown2007, levin04}, the primary
goal is to increase the areal coverage of the image; a single image is
usually deemed sufficient to represent the scene in a given region,
and the challenge is to register and blend the images to produce
seamless mosaics.  In \emph{photo-tourism} \citep{snavely2006,
  agarwal2009}, the multiple images provide different viewpoints and
levels of detail; again, a single image is typically sufficient to
capture the required level of detail or dynamic range.

% mention that lucky imaging usually throws out most of the data, and
% faster readouts mean more read noise?

% HDR ref?

% images -> scenes?

% mention systematics?

Our problem setting also differs from typical astronomical image
processing for scientific purposes.  Usually, great care is taken to
calibrate the detector so that pixel values can be mapped back to
intensity (photon counts) linearly.  Combining multiple images is then
a straightforward matter of registering the images and producing a
weighted sum.  At the faint end, the sensitivity of typical
astronomical images is limited by noise from reading out the detector
and from atmospheric and other background emission (with Poisson
noise) competing with the signal of interest.  Increasing the exposure
time thus increases the signal-to-noise ratio.  The Hubble Ultra-Deep
Field processed images, for example, were produced by summing hundreds
of exposures in each of four bands in order to increase the
sensitivity (depth) to detect extremely faint objects.  In our problem
setting, we cannot assume linearly calibrated images, so a more
elaborate method of combining images is required.

More generally, this project connects to other comprehensive projects
that hope to make use of large, heterogeneous collections of data:
Data can only be used simultaneously if it can be calibrated onto some
kind of common reference system.  The rank statistics used here
provide a robust basis for this.  More specifically, in astrophysics
there are no agreed-upon or generally useful robust methods for
combining heterogeneously processed images from disparate sources.
Images from multiple telescopes are rarely combined, even when
calibration is well understood; essentially never when it is ill
understood.  Historical astronomical image collections, for example,
contain sky coverage equivalent to the entire sky imaged hundreds or
even thousands of times over; yet they haven't been combined into any
kind of master image as of today.

\section{Method}

% also discuss how this is a method for aggregating votings for HUGE numbers of very long (millions) lists.

We have $N$ images $i$ that have been registered and resampled to the
same pixel grid.  (We will discuss \emph{how} we get these images
below.)  Each image contains pixel values (``data'') $d_i$, where
$d_i$ can be thought of as a vector as long as the number of pixels in
the image.  The intensity field $I$ (energy per time per solid angle
per logarithmic wavelength interval) generates the data $d_i$ only
after convolution with the appropriate point-spread function,
transformation to the relevant pixel grid, and sampling.  We represent
this transformation as a linear operator $H_i$ acting on the intensity
field.  The linearly transformed intensity field $H_i\cdot I$ is
transformed non-linearly through a monotonically increasing function
$f_i(\cdot)$ to ``data values'' and noise $e_i$ is added to make the
observed data:
\begin{eqnarray}
d_i &=& f_i(H_i\cdot I) + e_i
\end{eqnarray}
In reality, the noise is probably not precisely additive.  The
non-linear mapping $f_i$ includes detector effects such as saturation,
as well as post-processing (gamma correction, white balance, etc),
performed either in the camera or in photo-editing software.  We
assume that $f_i$ is monotonically increasing; If we don't have
further information about $f_i$, the absolute values of $d_i$ do not
carry useful information.  However, \emph{ranks} among observed values
can be exploited.

Combining ranks from a set of multiple images is a problem of
\emph{rank aggregation}. This problem is usually formalized as finding
a permutation which minimizes the Kendall-tau distance to the input
rankings (Kemeny rank aggregation), in which case it is known to be
NP-hard (for an overview, see \citet{Schalekamp}). Most methods either
specialize on aggregating many short ranked lists (e.g., for voting),
or few long lists (e.g., for meta-search engines, or certain
bioinformatics applications). One way to build efficient heuristic
methods is to use a \emph{positional} approach, i.e., to seek a
permutation in which the position of each element is ``close to the
average position'' of the element in the input lists
\citep{Schalekamp}. We will propose such a method for our problem,
fully aware that this does not solve the exact rank aggregation
problem. Indeed, for the setting of few long lists, a common approach
is to use Markov chain based heuristics (for a discussion, see
\citet{Lin}). Our setting contains long lists (with millions of
entries) \emph{and} a potentially large number of such lists.

% Moreover, our setting is nontrivial in that we aggregate partial
% ranks, since our subimages may cover different parts of the sky.

Let \red{$c$ be the \emph{consensus} image} that we are trying to
compute, containing $P$ pixels. This may be a rectangular image of a
certain area of the sky, or an array of possibly non-rectangular
pixels uniformly covering the celestial sphere.  Since we are only
concerned with pixel ranks, we initialize it as a vector whose
elements form a random permutation of $\{1,\dots,P\}$. Our algorithm
will ensure that after each update step, the entries of $c$ are again
a permutation of $\{1,\dots,P\}$, aggregating the information about
the pixel ranks as presented in the \red{\emph{observed images} or
  \emph{data} $d_i$}. The data will usually cover a proper
sub-field of $c$ only, modeled by a binary \red{\emph{mask} vector
  $m_i$} whose entry is $1$ if and only if the corresponding entry is
covered by $d_i$.  The data image $d_i$ will update the consensus
image $c$ only according to its footprint, i.e., in the
\red{restriction of $c$ to the mask $m_i$, denoted $c|_{m_i}$} (i.e.,
the sub-image consisting of those pixels where $m_i=1$).

\codefigure

Along with the consensus image $c$, we keep a vector of
\red{\emph{votes} $v$}, initialized to $(0,\dots,0)\in {\bf R}^p$. For
each pixel of $c$, the corresponding value in $v$ records how many
data images have contributed to it so far by voting for its rank.

Let {\tt argsort} be an $O(n \log n)$ function that, given a vector
$d$ of length $n$, returns a permutation $\sigma$ so that $d(\sigma)$
is sorted in non-decreasing order.\footnote{$d(\sigma)$ is a
  composition of $\sigma$ and $d$ where the vector $d$ is thought of
  as a function on its index set} The inverse permutation
$\sigma^{-1}$ is a function that maps an index $p$ in the vector $d$
to the \emph{rank} of its entry $d_p$ among the set of all entries.

The proposed \Enhance\ algorithm proceeds as shown in
\figref{fig:code}.  Let us discuss some details and properties of the
method.
\paragraph{Tied ranks.}
 Due to clipping and quantization, a data image $d_i$ often has sets
 of identical entries. In this case, there are several permutations
 $\tau$ sorting it. In practice, we average the corresponding ranks
 $\tau^{-1}$ to get what is called the \emph{tied rank}; with some
 abuse of notation, we refer to it as $\tau^{-1}$.

\paragraph{Invariant Histogram.}
Since the pixel values in the consensus image are only permuted, the
final image has the same pixel histogram as its initialization.  It
can easily be post-processed to make a pixel histogram that is in
accord with prior expectations, for instance taking into account
models of sensor noise, sky background, star brightness distributions,
point spread functions, etc; or histogram-matched to a given image.

 Above, we initialized $c$ to a random permutation of
 $\{1,\dots,P\}$. Other initializations are possible as well,
 including
\begin{itemize}
\item a random image with a specified histogram (which will remain
  invariant, since the entries of $c$ only get permuted by the
  algorithm),
\item a `current best guess' image, for instance based on existing sky
  maps, or a previous run of the algorithm.
\end{itemize}

\paragraph{Convergence.}
The choice of the voting weights ensures that for each pixel, each
data image that contributed to it will have had the same relative
weight in the algorithm's averaging step. This does not, however,
imply that the algorithm is invariant to the ordering of the images:
As images are combined, the total weight vector $\tilde{v}$
accumulates.  Eventually, the total weight becomes large enough that
new images simply do not have enough influence to change the ordering
of the pixels in the consensus image. The maximal image size is $P$
pixels, hence the maximal rank difference that can be observed in an
image $d_i$ is $P-1$; this ranking will come with a weight of $1$. The
current consensus image, however, will receive weight $v$. Once all
entries of $v$ are at least $P$, i.e., once every pixel has been
covered by at least $P$ data images, the averaging step can no
longer change the ranking of the consensus image, and the algorithm
has converged, at the expense of any influence of late-arriving input
images.

\examplefigure

In our application, we are far from this happening ($P$ is of order
$10^6$, and we are nowhere near to having a comparable number of
images at this point). Nevertheless, one can think of strategies to
avoid this:
\begin{itemize}
\item Use mini-batches of image that are combined to a separate
  consensus image with a nontrivial voting vector, which can then be
  combined with the current consensus image (using a straightforward
  generalization of our voting formula).
\item Assign non-binary weights to data images, to ensure that
  rankings are considered more reliable where the pixel values were
  different, and the rankings thus reliable. For instance, if 10\% of
  the pixels have exactly the same value (due to discretization or
  clamping, say), then the ranks of those pixels are uncertain at
  $\sim 10\%$; if we assume a constant per-pixel error standard
  deviation in the input images, then the rank error is the per-pixel
  error times the derivative of the ranking function, ie, the
  normalized histogram value.  Given Gaussian noise this would suggest
  a per-pixel weighting of ${1/h(d_i)}^2$, where $h$ is the histogram.
  We have not explored this option in this paper but expect it would
  improve our results by increasing the weights of images with high
  signal-to-noise.
\end{itemize}

%  This happens---input images lose any capability for reordering
%  pixels---when (or really before) $i > \mbox{\tt length}(\rho)$.

\paragraph{Astrometry.net}
Before combining the images, we need to recognize them (as images of
the night sky) and register them to a common celestial coordinate
system.  These tasks are performed by a system, \An, that has been
described in the astronomical literature \citep{lang2010}, but not in
the machine learning or statistics literature.  \An\ image recognition
and calibration proceeds in three stages:

In the star-measuring phase, a rough noise estimate is made in the
input image by a median-absolute-difference analysis and statistically
significant peaks are identified as possible ``stars''.  Each star is
given a measured centroid and brightness by a fit of a quadratic
function to a pixel patch centered on its peak pixel in the image.

In the geometric-hashing phase, sets of four stars are considered
sequentially (and exhaustively in an ordering based on star
brightnesses, limited only by CPU time).  For each set of four stars,
a geometric hash is computed that is invariant to translation,
rotation, and scale.  This continuous, four-dimensional hash code is
looked up in an index of hashes of known four-star asterisms
(combinations of stars), with a finite tolerance to account for
measurement noise, small camera distortions, and stellar proper
motions.  Any match of a four-star hash with an indexed hash creates a
``proposal'' for the mapping between image coordinates and celestial
coordinates.  The pre-computed index of four-star hashes is designed
to cover the entire sky densely, not over-use individual stars (which
might be noisy or variable or wrong in the input catalogs), and make
use of stars that are likely to be co-visible in normal images.

In the decision phase, every proposal about the coordinate mapping
generated in the hashing phase is used to make predictions about the
positions of \emph{other} stars in the image (other than the four used
to make the hash), based on the positions of known stars.  A robust
likelihood for the proposal is computed, permitting there to be known
stars not detected in the input image, and detected stars in the input
image that are not known.  The posterior probability for the proposal
and a null hypothesis (that the hash alignment has been found by
chance) are compared, to make a decision in the context of Bayesian
decision theory with a well-defined utility function.  The utility
function is designed to be conservative (that is, reject a match
unless there is overwhelming support for it).

\histeqfigure

After a finite amount of CPU time, the system either returns a failure
or else the full mapping between image coordinates and celestial
coordinates.  This mapping is delivered in standards-compliant
astronomical format (FITS WCS).  In detail, it delivers a full
polynomial model of the camera distortions away from a tangent-plane
projection.  In particular, for the purposes of this project, the
output of \An\ includes the celestial footprint of the image and the
precise transformation from pixel position to celestial position for
every pixel in the input image, thus solving the problem of
registering the images to a common pixel grid.

\section{Experiments}

Our method is defined for collections of images of a fixed or
near-fixed scene from a near-fixed viewpoint.  The images must be
registered and resampled onto a common pixel grid, with associated
mask images showing the registered image footprint on the reference
``canvas''.  We obtained two example image collections of extended
astronomical objects from the Web, one for the galaxy NGC~5907 and one
for the interacting pair of galaxies Messier 51a and 51b.  In the case
of NGC~5907, the goal was to use (shallow, non-scientific) images
available on the Web to rediscover or confirm astrophysically
important but very faint (low surface-brightness)
features discovered in the outskirts of the galaxy \citep{dmd2008}.
We chose the first target (NGC~5907) since a high-quality ``ground
truth'' image is available, and the second target (Messier 51) because
we knew a large number of images would be available.

\ngcresultsfigure

We generated the image collections using the ``image search'' APIs of
\flickr, \corp{Bing}, and \corp{Google}.  For the first target, the
galaxy NGC~5907, we used the search terms ``NGC5907'' and
``NGC~5907''.  We retrieved as many results as the APIs would allow
(4000 for \flickr, 1000 for \corp{Bing}, and 100 for \corp{Google}),
and repeated the search with different filters on the image size in
order to get the largest possible set of results.  Examples of the
images produced by this search are shown in \figref{fig:examples}.

Many images returned by the Web search services were not images of the
sky, or else not useful images, or else images of the sky but not
including the target; attempted recognition with the \An\ system
filtered these out.  In \figref{fig:examples} we have marked the
images that were recognized as images of the sky.  For this
experiment, we also filtered the images by hand to remove images with
very prominent labeling or overplotted diagrams or text decoration,
and removed the highest-quality images (taken by \citet{dmd2008}, or
derived or reprocessed from those images) that individually show the
low intensity features of greatest interest.  We believe both removals
could have been performed automatically with objective operations on
the pairwise rank statistics, but implementation was beyond the
deadline-limited scope of this project.

The Web search services returned a total of 2034 image URLs, of which
1967 were retrievable and had unique contents, and 1817 were jpeg
images.  Of these, 405 were recognized as images of the night sky
overlapping our target region near NGC~5907.  Of these, 48 were images
from \citet{dmd2008} or reprocessings of those images, and we marked
an additional 59 images as having excessive annotations or markings.
The experiments below use the remaining 298 pixel-aligned images of
NGC~5907.

A reference $900 \times 900$ pixel grid (canvas) in a tangent-plane
projection of the sky $30 \times 30$~arcmin$^2$ in solid angle
(roughly the area of the full moon) was generated, centered on the
fiducial celestial position of
\break
%
%\setlength{\parskip}{0pt}
%NGC 5907 (RA, Dec = 229, 56.3 degrees J2000).  The \break

\clearpage
\mfiftyonefigure

\noindent
NGC 5907 (RA, Dec = 229, 56.3 degrees J2000).  The 
\An-recognized images---which in recognition are also
given standards-compliant celestial-coordinate-system meta-data---were
resampled by nearest-neighbor resampling onto the reference grid.
Many of the input images did not fully cover the reference image
canvas; we produced a mask indicating which reference pixels were
covered by each input image.
\setlength{\parskip}{0.5pc}

%\clearpage
%\mfiftyonefigure

For the first experiment with the NGC~5907 data, we chose four images
with different dynamic range (shown in \figref{fig:four}).
We performed a weighted average of these images, weighted by their
mask vectors.  We also combined them according to the
\Enhance\ algorithm given in the Method Section.  The reference image
was initialized with the histogram-equalized average image.  The
results are shown in \figref{fig:four}.  \Enhance\ combines
the images to produce an image with high overall dynamic range.

For the second experiment, we ran the \Enhance\ algorithm on all 298
images in the collection, initializing with the mean image.  We ran
the red, green, and blue color channels separately.  The results are
shown in \figref{fig:5907}.

We compute an approximate measure of the Kendall tau rank-correlation
coefficient between the \Enhance\ consensus image and one of the
high-quality deep images of NGC~5907 that we removed from the input
image set (with the manual filtering).  We compare this to the Kendall
tau for a simple mask-weighted average of the input images and the
same deep image.  The \Enhance\ output gets Kendall tau of 0.355 and
the weighted average 0.183, demonstrating that \Enhance\ produces a
consensus image much closer to our proxy for ``ground truth''.
Furthermore, the mean inter-image Kendall tau between pairs of input
images is 0.165 whereas the mean Kendall tau between input images and
the consensus image produced by \Enhance\ is 0.249, demonstrating that
the consensus is good.

For the third experiment, the Messier 51 image collection was used.
Analysis was identical, except that the search terms included several
aliases for the galaxy: ``M 51'', ``Messier 51'', ``Whirlpool
Galaxy'', ``M 51a'', ``M 51b'', ``NGC 5195'', ``NGC 5194'', ``UGC
8494'', ``PGC 4741'', ``Arp 85'', plus the same phrases without
spaces.  The reference grid was centered on RA,Dec = (202.5, 47.22)
degrees J2000, with $1080 \times 1080$ pixels and an area of $18
\times 18$~arcmin$^2$.  The search yielded 13,472 URLs, of which
12,793 were unique retrievable files, 11,709 were jpeg images, 2072
were recognized by \An, and 2066 overlapped our region of interest.
(We instructed \An\ to search only in a small region around M51.)  The
results, which took about 40 minutes on a single processor, are shown
in \figref{fig:m51}.

% Finally, we consider the possibility that images can be ordered or chosen
% according to their information content or value for ranking pixels.
% Kendall tau, performance, examples, whatever... DSTN.
% Random sparse approximation to Kendall tau?
% Can we get a better stack by excluding bad-tau images?
% The results are shown in \figurename~\ref{fig:kendall}
% \begin{figure}[ht]
% ~\hfill DSTN GOTCHU KENDALL STATS \hfill~
% \caption{\textsl{top-left:} Example of a pair of images with very high Kendall tau.
%   Note that they come from identical source data.
%   \textsl{top-right:} Example of a pair of images with very different Kendall tau.
%   Note that one image is almost uninformative.
%   \textsl{bottom:} The NGC~4907 \Enhance\ image created with a subset of the full image collection.
%   The subset was created by dropping images with low Kendall tau against the \Enhance\ image shown in \figurename~\ref{fig:5907}.
%   \label{fig:kendall}}
% \end{figure}

\section{Discussion}

We have proposed a system that can automatically combine uncalibrated,
processed night-sky images to produce high quality, high dynamic-range
images covering wider fields.  The consensus images output by
\Enhance\ shown in \figurename s~\ref{fig:5907} and \ref{fig:m51}
reveal very low-surface brightness tidally disrupted features from
violent galaxy--galaxy interactions between subcomponents of each
system.  These faint features are of great astrophysical
importance---they permit (in principle) measurement of the ages, mass
ratios, and orbital configurations of recent merger events---and yet
some of them are not clearly visible in any of the input images.
\Enhance\ creates opportunities for new astronomical discoveries.

Our method scales linearly in the number of images, making it
applicable to the large (and increasing) number of astronomical images
available. For each update step, the time complexity is $n \log n$
(due to the sorting algorithm) in the image size $n$ of the new image
being added (and independent of the consensus image), taking a few
seconds on a laptop for a 3 Megapixel image in matlab. The method can
thus be applied to internet-scale problems.

We have set up a Web site that allows users to submit images for
combination into an Open-Source Sky Map.\footnote{%
  \url{http://nova.astrometry.net}}
The systems keeps an overall sky map (in
celestial coordinates), whose brightness ranks are initialized as a
random permutation, as described above. Every time an image is
submitted, the system recognizes and calibrates the image with
\An\ and merges its rank information into our overall rank map as
described.  We make use of the astronomical-standard HealPix
pixelization of the sky for the all-sky map \citep{healpix}.  In return,
users receive annotated versions of their images, as well as a
histogram-matched version of the corresponding part of the current
state of the map. The former allows the identification of known
astronomical objects, while the latter is informative with respect to
any differences to the present images, helpful for identifying
transient events as well as noticing any issues with calibration or
instrumentation.  A limitation for these applications of the current
system, and a promising direction for future research, is the
inclusion of point spread function modeling.

One key value of pixel rank representations of images is that they
permit identification of non-identical input images that nonetheless
were created from the same original source data (telescope or camera
images).  In principle we should identify these duplicates and use
only one copy.  However, the inclusion of duplicates or images related
by having common raw-data origins does not have a large negative
impact on the \Enhance\ output; these images just end up having higher
overall influence on the consensus ranks in the final output.

The specific \Enhance\ algorithm given here works on static
two-dimensional scenes.  That is, it cannot be applied to
time-variable scenes or three-dimensional scenes with multiple camera
viewpoints without substantial modification.  However, there are many
natural-image applications for which the assumptions of
\Enhance\ apply.  Shots of famous landmarks from other famous
landmarks could be used (for example the Brooklyn Bridge viewed from
the Empire State Building Observation Deck).  Images taken in
different weather conditions with different cameras, processed
differently could be registered and then combined with \Enhance.

Finally, one scientifically very important application for
\Enhance\ is in the analysis of historical photographic plate data.
Astronomical data in photographic (glass) plate form dates back to the
1880s or earlier, and the Harvard Plate Archives alone contain half a
million plates, covering the sky hundreds of times over.  Different
emulsions, different exposures, different developing, and different
pre-exposure plate treatment (``hypering'') lead to different
non-linear responses of emulsion density to incident intensity.  These
valuable plate images can be compared and combined with pixel-rank
statistics without a full treatment of all nonlinear effects.
\Enhance\ could have a large impact on historical astronomy and
astrophotography.

%\clearpage

\newcommand{\AJ}{The Astronomical Journal}
\newcommand{\aj}{\AJ}
\newcommand{\apj}{The Astrophysical Journal}
\newcommand{\aap}{Astronomy \& Astrophysics}
\newcommand{\ICCP}{International Conference on Computational Photography}

\bibliography{bs}{}
\bibliographystyle{plainnat}

\end{document}